\documentclass[sigconf]{acmart}
\usepackage{booktabs}
\usepackage{multirow}
\usepackage{subcaption}
\usepackage{tcolorbox}
\usepackage{adjustbox}
\usepackage{xspace}
\usepackage{threeparttable}
\AtBeginDocument{%
  }
\setcopyright{acmlicensed}
\copyrightyear{2018}
\acmYear{2018}
\acmDOI{XXXXXXX.XXXXXXX}
\acmConference[Conference acronym 'XX]{Make sure to enter the correct
  conference title from your rights confirmation email}{June 03--05,
  2018}{Woodstock, NY}
\acmISBN{978-1-4503-XXXX-X/2018/06}

\usepackage{gradient-text}
\newcommand{\ama}{\gradientRGB{AMA}{0,0,185}{240,0,15}\xspace}

\begin{document}


\title{When Agents Trade: Live Multi-Market Trading Benchmark for LLM Agents}

\author{Lingfei Qian}
\affiliation{%
  \institution{\normalsize The Fin AI}
  \city{\normalsize New Haven}
  \state{\normalsize Connecticut}
  \country{\normalsize USA}
}
\author{Xueqing Peng}\affiliation{%
  \institution{\normalsize The Fin AI}
  \city{\normalsize New Haven}
  \state{\normalsize Connecticut}
  \country{\normalsize USA}
}
\author{Yan Wang}
\affiliation{%
  \institution{\normalsize The Fin AI}
  \city{\normalsize New Haven}
  \state{\normalsize Connecticut}
  \country{\normalsize USA}
}
\author{Vincent Jim Zhang}
\affiliation{%
  \institution{\normalsize The Fin AI}
  \city{\normalsize New Haven}
  \state{\normalsize Connecticut}
  \country{\normalsize USA}
}
\author{Huan He}
\affiliation{%
  \institution{\normalsize The Fin AI}
  \city{\normalsize New Haven}
  \state{\normalsize Connecticut}
  \country{\normalsize USA}
}

\author{Zehan Li}
\affiliation{%
  \institution{DeepKin}
  \city{\normalsize New Haven}
  \state{\normalsize Connecticut}
  \country{\normalsize USA}
}

\author{Hanley Smith}
\affiliation{%
  \institution{PAAL AI}
  \city{\normalsize Toronto}
  \state{\normalsize Ontario}
  \country{\normalsize Canada}
}

\author{Yi Han}
\affiliation{%
  \institution{Georgia Institute of Technology}
  \city{\normalsize Atlanta}
  \state{\normalsize Georgia}
  \country{\normalsize USA}
}

\author{Yueru He}
\affiliation{%
  \institution{Columbia University}
  \city{\normalsize New York}
  \state{\normalsize New York}
  \country{USA}
}

\author{Haohang Li}
\affiliation{%
  \institution{Stevens Institute of Technology}
   \city{\normalsize Hoboken}
  \state{\normalsize New Jersey}
  \country{\normalsize USA}
}
\author{Yupeng Cao}
\affiliation{%
  \institution{Stevens Institute of Technology}
   \city{\normalsize Hoboken}
  \state{\normalsize New Jersey}
  \country{\normalsize USA}
}
\author{Yangyang Yu}
\affiliation{%
  \institution{Stevens Institute of Technology}
   \city{\normalsize Hoboken}
  \state{\normalsize New Jersey}
  \country{\normalsize USA}
}

\author{Alejandro Lopez-Lira}
\affiliation{%
  \institution{University of Florida}
  \city{\normalsize Gainesville}
  \state{\normalsize Florida}
  \country{\normalsize USA}
}

\author{Peng Lu}
\affiliation{%
  \institution{Université de Montréal}
  \city{\normalsize Montréal}
  \state{\normalsize Quebec}
  \country{\normalsize Canada}
}
\author{Jian-Yun Nie}
\affiliation{%
  \institution{Université de Montréal}
  \city{\normalsize Montréal}
  \state{\normalsize Quebec}
  \country{\normalsize Canada}
}

\author{Guojun Xiong}
\affiliation{%
  \institution{Harvard University}
   \city{\normalsize Cambridge}
  \state{\normalsize Massachusetts}
  \country{\normalsize USA}
}

\author{Jimin Huang}
\authornote{Corresponding Author}
\affiliation{%
  \institution{\normalsize The Fin AI}
   \city{\normalsize New Haven}
  \state{\normalsize Connecticut}
  \country{\normalsize USA}
}
\affiliation{%
  \institution{\normalsize National Centre for Text Mining, University of Manchester}
  \city{\normalsize Manchester}
  \country{\normalsize United Kingdom}
}

\author{Sophia Ananiadou}
\affiliation{%
  \institution{\normalsize National Centre for Text Mining, University of Manchester}
  \city{\normalsize Manchester}
  \country{\normalsize United Kingdom}
}

\renewcommand{\shortauthors}{Lingfei et al.}

\begin{abstract}
Although Large Language Model (LLM)-based agents are increasingly used in financial trading, it remains unclear whether they can reason and adapt in live markets, as most studies test models instead of agents, cover limited periods and assets, and rely on unverified data. To address these gaps, we introduce Agent Market Arena (\textsc{\ama}), the first lifelong, real-time benchmark for evaluating LLM-based trading agents across multiple markets. \textsc{\ama} integrates verified trading data, expert-checked news, and diverse agent architectures within a unified trading framework, enabling fair and continuous comparison under real conditions. It implements four agents, including InvestorAgent as a single-agent baseline, TradeAgent and HedgeFundAgent with different risk styles, and DeepFundAgent with memory-based reasoning, and evaluates them across GPT-4o, GPT-4.1, Claude-3.5-haiku, Claude-sonnet-4, and Gemini-2.0-flash. Live experiments on both cryptocurrency and stock markets demonstrate that agent frameworks display markedly distinct behavioral patterns, spanning from aggressive risk-taking to conservative decision-making, whereas model backbones contribute less to outcome variation. \textsc{\ama} thus establishes a foundation for rigorous, reproducible, and continuously evolving evaluation of financial reasoning and trading intelligence in LLM-based agents.
\end{abstract}




\received{20 February 2007}
\received[revised]{12 March 2009}
\received[accepted]{5 June 2009}

\maketitle

\section{Introduction}
\textit{Can large language model (LLM)-based agents truly trade in real time?}
This question sits at the cutting edge of financial artificial intelligence, where language models are transforming from static text generators into autonomous decision-making systems that perceive, plan, and act in dynamic environments~\cite{guo2024large,lopez2025can}. Financial markets, which are volatile, high-dimensional, and unforgiving, offer perhaps the most stringent test of this emerging intelligence~\cite{al2024stock,lin2024stock,pagliaro2025artificial}. Trading is not about isolated predictions, but about making sequential, high-stakes decisions in a world that changes every second. To succeed, agents must not only understand market narratives but also adapt continuously to unfolding events, uncertain feedback, and adversarial forces~\cite{deliu2024reinforcement,li2024developing,xiao2025retrieval}.

There are multiple efforts attempting to answer this question. DeepFund~\cite{li2025time} marks a significant step forward, introducing a multi-agent framework that integrates live data streams to simulate real-time decision-making. InvestorBench \cite{li2024investorbench} expanded static evaluation into trading simulations, replaying historical data to assess hypothetical returns. Beyond these, benchmarks focused on financial sentiment analysis \cite{malo2014good,nasiopoulos2025financial}, event extraction \cite{jacobs2020extracting,han2022duee}, and financial text understanding ~\cite{zhang2023fineval,xie2023pixiu} advanced domain comprehension but did not capture adaptivity or feedback, the hallmarks of real-world trading.



Despite all these efforts, it is still nontrivial to answer the question, as existing benchmarks continue to face three fundamental challenges. Firstly, existing work mainly \textbf{evaluates models rather than agents}. Frameworks such as DeepFund \cite{li2025time} and InvestorBench \cite{li2024investorbench} are tied to fixed agent structures, which evaluate model backbones rather than the agents themselves. Even when different LLMs are substituted, performance trends remain similar, suggesting that identical frameworks lead to convergent decision behaviors. Secondly, the \textbf{evaluation scope is extremely limited}. DeepFund tests agent with only a limited time period of 24 days that focuses only on U.S. stocks (Apple, American Express, Bank of America, Coca-Cola, and Chevron), meaning just 24 decisions per stock in total. Such a narrow setup offers little insight into how agents generalize across diverse market regimes, particularly when transitioning between bullish and bearish conditions, or longer trading horizons. 
Thirdly, the input \textbf{data lacks consistency and verification}. Existing trading benchmarks rely on heterogeneous data sources, such as \texttt{yfinance}\footnote{\url{https://pypi.org/project/yfinance/}}
 and \texttt{Finnhub}\footnote{\url{https://finnhub.io/}}, to
 retrieve market prices and news. However, existing benchmarks do not follow a unified data acquisition protocol, and their news feeds frequently overlap or contradict caused by leveraging multi API sources. The same market event may appear multiple times or with varying emphasis across sources, introducing redundancy and inconsistency. Such unverified and noisy inputs distort the informational environment available to agents, thereby undermining the reliability and interpretability of their trading decisions.

To fill this gap, we propose Agent Market Arena (\textsc{\ama}), the first lifelong, real-time, and multi-class-asset evaluation framework for LLM-based trading agents, built entirely on verified and continuously updated market data. 
\textsc{\ama} has been live for two months and will remain active as an open, evolving benchmark that grows alongside real markets.
The defining features of \textsc{\ama} lie in its unified protocol, verified information streams, and multi-class-asset market coverage. The protocol standardizes how agents interact with the market. Every agent starts with the same initial capital, trades at a fixed daily time, and follows identical execution and feedback rules, ensuring fair and reproducible evaluation.
\ama also maintains a verified information pipeline that continuously integrates price data, financial news, and company reports. Each stream is standardized, summarized, and checked for factual accuracy, consistency, and neutrality through a expert verification, providing reliable and unbiased inputs for trading decisions.
In addition, \textsc{\ama} connects to diverse markets spanning equities and cryptocurrencies, offering a realistic and dynamic testbed for adaptive reasoning.

Based on it, for the first time, a set of state-of-the-art LLM–based agent systems has been deployed to operate continuously in two stocks (\textsc{Tesla} (TSLA) and \textsc{Biomarin Pharmaceutical} (BMRN)) and two cryptocurrencies (\textsc{Ethereum} (ETH) and \textsc{Bitcoin} (BTC)). The deployment began two months ago and has since run without interruption, executing daily trades and adapting to real-time market fluctuations. The system includes \textbf{InvestorBench}~\cite{li2024investorbench}, \textbf{TradeAgent}~\cite{xiao2024tradingagents}, \textbf{HedgeFundAgent}~\cite{virattt_ai-hedge-fund}, and \textbf{DeepFundAgent}~\cite{li2025time}, each representing distinct reasoning strategies, ranging from aggressive trading to conservative hedging, and from role-based to memory-adaptive decision-making. 
For each agent framework, we further test multiple LLM backbones, enabling \textsc{\ama} to disentangle the effects of agent architecture and model choice under identical trading conditions. All trading activities are logged through the \textsc{\ama} evaluation pipeline\footnote{https://ama.thefin.ai/live}, producing a continuously growing dataset of real-world decisions and outcomes. Performance is measured using standard financial metrics, including cumulative return, annualized volatility, maximum drawdown, and sharpe ratio, allowing consistent and interpretable comparison across agents and markets.

Our two-month live trading results show that LLM-based agents can consistently show promising capabilities in making trading decisions, demonstrating genuine reasoning ability in dynamic financial environments. Across all assets and sessions, agent architecture proved to be the dominant factor shaping behavior. InvestorAgent achieved Sharpe ratios of 6.47 on TSLA, outperforming Buy \& Hold with smoother returns and lower drawdowns. DeepFundAgent reached 2.45 on BTC, showing that memory-based reasoning improves adaptability under volatility. TradeAgent and HedgeFundAgent favored aggressive risk selection, delivering remarkable returns accompanied by elevated volatility.
Changing the LLM backbone among GPT-4o\cite{hurst2024gpt}, 
GPT-4.1\cite{OpenAI2025GPT4_1},
Claude-3.5-haiku\cite{anthropic2025claudeHaiku}, 
Claude-sonnet-4\cite{anthropic2025claude4}, 
and Gemini-2.0-flash\cite{GoogleCloudPlatform2025introGemini} shifted outcomes by less than agent switching, confirming that these patterns are stable and reproducible across models.


Our contributions could be summarized as follows.

$\triangleright$Firstly, we introduce Agent Market Arena,
the first lifelong, real-time benchmark for evaluating LLM-based financial agents under live market conditions.

$\triangleright$Secondly, we construct a verified dataset that covers multi markets, integrating real trading data with expert-reviewed financial news to ensure factual accuracy, neutrality, and continuous updates.

$\triangleright$
Thirdly, we implement a diverse suite of trading agents, including single-agent, multi-agent, with various role and memory-adaptive designs, to examine how agent structures and the trading style shape performance across assets. 

$\triangleright$
Finally, we build a transparent leaderboard that records profitability and adaptability over time, offering an open and reproducible platform for studying intelligent financial agents.

\section{Related Work}
\subsection{Benchmarking LLMs in Financial Domain}
There is now a growing body of work on financial benchmarks. Early financial NLP benchmarks~\cite{loukas2022finer,sharma2022finred,xie2023pixiu,chen2021finqa,chen2022convfinqa,wang2025fintagging} such as FinBen~\cite{xie2024finben} expanded evaluation beyond sentiment and QA to include retrieval augmented tasks and multi‑step decision-making problems, yet they remained largely static and English‑centric. There were also efforts~\cite{peng2025plutus,zhang2024dolares} introduced benchmarks broadening the linguistic focus beyond English and Chinese, while MultiFinBen\cite{peng2025multifinben} extended these observations to multilingual and multimodal evaluation. FinReason\cite{qian2025fino1} further introduced financial reasoning tasks, showing that generic reasoning models degrade on financial data.

\subsection{Trading Agents and Live Benchmark}
There is also a growing line of research that explores using LLMs or agents to make decisions and study their performance in decision-making. A growing set of studies has begun to design and analyze trading agents. FinMem\cite{yu2025finmem} and InvestorBench\cite{li2024investorbench} introduced memory‑augmented LLM agents and benchmarked them over backtesting market information. FLAG‑TRADER\cite{xiong2025flag} combined language models with policy‑gradient training for sequential decision making. FinCon\cite{yu2024fincon} extended the memory updating to multi‑agent interaction, proposing a framework for single‑stock trading and portfolio management, incorporating hierarchical communication and risk control. HedgeFundAgents\cite{virattt_ai-hedge-fund} advanced this direction by modeling a hedge‑fund hierarchy, coordinating a central manager with hedging experts to achieve more robust returns. TradeAgents\cite{xiao2024tradingagents} further explored decentralized coordination among specialized trading agents, highlighting how division of labor can improve market decision making. Yet these systems often operate in static environments. To address benchmark leakage and excessive intervention, DeepFund\cite{li2025time} proposed a live multi‑agent arena for fund investment, enabling dynamic evaluation of LLM‑driven strategies.


\section{Agent Market Arena: Real-time trading and competition}

\begin{figure}
    \centering
    \includegraphics[width=\linewidth]{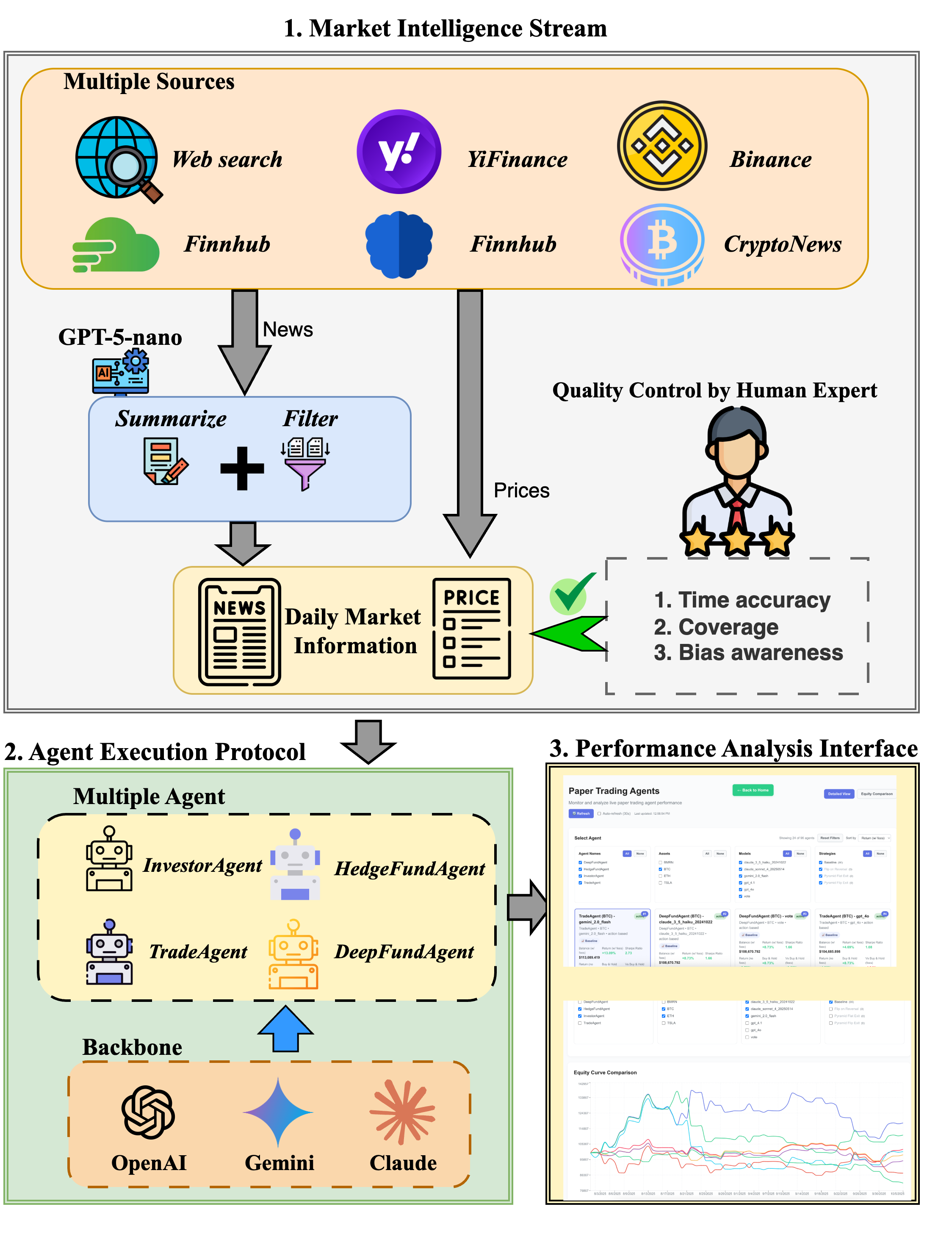}
    \caption{Overall framework of Agent Market Arena.}
    \label{fig:framework}
\end{figure}

The overall framework of Agent Market Arena is illustrated in Figure ~\ref{fig:framework}.
It is designed to evaluate how different agent frameworks perform when given access to the real-time market environment. 

It consists of three main components. 
\textbf{First}, the \textit{Market Intelligence Stream (MIS)} continuously aggregates data from multiple sources and generates high-quality daily summaries of market conditions. The summary is reviewed by financial experts to ensure accuracy, completeness, and neutrality.
\textbf{Second}, the \textit{Agent Execution Protocol (AEP)} provides a unified environment for deploying multiple trading agents, allowing each to access the verified updates and make independent trading decisions across evaluated assets.
\textbf{Finally}, the \textit{Performance Analytics Interface (PAI)} offers a continuously updated dashboard that tracks and compares the real-time performance of all agents built on different LLM backbones, providing a transparent view of profitability, stability, and adaptability.

\subsection{Market Intelligence Stream (MIS)}
We begin by collecting real-time market data from a wide range of heterogeneous sources to ensure both coverage and accuracy. These include general-purpose APIs and online news aggregators. We leverage sources such as OpenAI Web Search~\footnote{https://platform.openai.com/docs/guides/tools-web-search?api-mode=responses}, Finnhub, NewsData\footnote{https://newsdata.io/}, yifinance, CryptoNews\footnote{https://cryptonews.com/}, and Binance\footnote{https://www.binance.com/en}, which provide access not only to news streams from established financial outlets, but also to social media content. This ensures that both API-driven feeds, direct news content, and integrated social media signals from platforms such as Twitter\footnote{https://x.com/?lang=en} and Reddit\footnote{https://www.reddit.com} are incorporated to capture emerging sentiment. 
This multi-source integration reduces the bias of relying on a single channel and provides a broader, more balanced, and truly real-time view of the market environment.

Despite the diversity of sources, much of the content may partially overlap, which could introduce bias if the same information repeatedly appears to the agents. To address this, we apply a summarization step to consolidate the news and prevent redundancy from distorting the evaluation. We use GPT-5-nano for this task, with implementation details provided in Appendix~\ref{quality}.

\textbf{Quality control.} To rigorously evaluate the quality of the summarized news, we randomly sample 20 daily items from the generated outputs, covering both equities and cryptocurrencies, 20 days for each. 
Following the focus of previous works, two independent annotators assess each item according to the following criteria:
(1) \textbf{Date Accuracy}\cite{dougal2012journalists}: Each item must correspond precisely to the intended trading date and the associated assets.
(2) \textbf{Coverage}\cite{tetlock2007giving}: The sample must include all major news of the day directly related to the target assets across the principal categories that drive prices, ensuring that no significant asset-related information is omitted.
(3) \textbf{Bias Awareness}\cite{gentzkow2006media}: Determine whether the summarization process introduces any additional or noticeable bias that is absent from the original content.
For each criterion, we adopt a three-level scale: 0 indicates not satisfied, 1 indicates partially satisfied, and 2 indicates fully satisfied. 

The evaluation was conducted independently by two annotators using the 20-day samples for both equities and cryptocurrencies.
Results show strong consistency: date accuracy agreement reached 87.5\%, coverage 92.5\%, and bias awareness 100\%. Most date discrepancies (7 cases, score = 1) were due to API latency, where the retrieved link was current but the article content lagged by one or two days. Four cases received a score of 1 for coverage, primarily due to partially incomplete summaries that omitted minor updates—such as relevant company name changes, indicating shifts in business direction. These omissions were considered acceptable, as all major market-moving events were accurately captured within the evaluated summaries.
For the bias awareness, no summarization was found to introduce bias or sentiment beyond what existed in the original news sources.
This rigorous daily quality control ensures that the benchmark reflects high-fidelity, real-time market narratives and minimizes the propagation of outdated or biased information into agent evaluation.

\subsection{Agent Execution Protocol (AEP)}
The \textit{Agent Execution Protocol (AEP)} defines how all agents in \textsc{\ama} interact with the market environment. It standardizes three core components, including inputs, outputs, and configuration, to ensure that performance differences arise from reasoning ability rather than implementation variance.

\textbf{Inputs.}
Each agent receives identical daily inputs, including verified market data and the summarized news from the MIS. The information is structured into a unified prompt format containing asset identifiers, historical prices, recent news summaries, and contextual trading metadata. This consistent input design guarantees that all agents operate under the same market knowledge and temporal conditions.

\textbf{Outputs.}
Agents produce discrete daily actions from the same action space: \texttt{{BUY, SELL, HOLD}}. These actions are used to update the simulated portfolio and compute next-day returns. A \texttt{BUY} action means the agent expects the asset’s value to rise and increases its position; a \texttt{SELL} action indicates the opposite expectation; and a \texttt{HOLD} action reflects neutrality or uncertainty, keeping the current position unchanged. All trading decisions are executed synchronously at the same fixed time, ensuring fair comparison across models and markets.

\textbf{Prompts and Hyperparameters.}
Each agent is additionally guided by a structured system prompt that defines trading strategy and action descriptions. To maintain fairness, generation parameters such as temperature, retry times are fixed across all agents and backbones~\footnote{For more details, please check Appendix~\ref{parameter}}.

\textbf{InvestorAgent}\cite{li2024investorbench} is a single-agent framework that integrates a memory module to store historical trading decisions and contextual information. It analyzes past outcomes to extract actionable insights, which are then used to refine future optimization and decision-making processes.

\textbf{TradeAgent}\cite{xiao2024tradingagents} follows a multi-agent architecture inspired by real-world trading firms. Rather than relying solely on fixed technical indicators, it orchestrates specialized roles—such as fundamental, sentiment, news, and technical analysts—whose insights are debated by researcher agents and then aggregated by a trader agent. Final trade proposals are checked by risk management and a portfolio manager before execution.

\textbf{HedgefundAgent}\cite{virattt_ai-hedge-fund} models a hedge-fund team by orchestrating specialized agents that follow characters of famous trading people, like Aswath Damodaran and Ben Graham, among others, each simulate distinct roles in the investment process. Their signals are aggregated by a portfolio manager to produce trade intents, while a risk manager computes risk metrics and enforces position limits. 

\textbf{DeepFundAgent}\cite{li2025time} is designed to test adaptability by processing streaming inputs, generating trade signals, and updating positions based on history portfolio updates that stored in memories, as well as the media, technical analysis. This setup enables an assessment of how LLM-based decision systems perform in dynamic environments compared to rule-based or static benchmarks.


\begin{table*}[!htbp]
\centering
\small
\setlength{\textfloatsep}{5pt}
\renewcommand{\arraystretch}{0.95}

\begin{threeparttable}
\begin{adjustbox}{width=\textwidth}
\begin{tabular}{@{}ll
rrrrr rrrrr rrrrr rrrrr@{}}
\toprule
\multirow{2}{*}{\rotatebox{80}{\textbf{Agent}}} &
\multirow{2}{*}{\rotatebox{80}{\textbf{Model}}} &
\multicolumn{5}{c}{\textbf{TSLA}} &
\multicolumn{5}{c}{\textbf{BMRN}} &
\multicolumn{5}{c}{\textbf{BTC}} &
\multicolumn{5}{c}{\textbf{ETH}} \\
\cmidrule(lr){3-7}\cmidrule(lr){8-12}\cmidrule(lr){13-17}\cmidrule(lr){18-22}
& & \rotatebox{80}{\textbf{CR\% (↑)}} & \rotatebox{80}{\textbf{AR\% (↑)}} & \rotatebox{80}{\textbf{AV\% (↓)}} & \rotatebox{80}{\textbf{SR (↑)}} & \rotatebox{80}{\textbf{MDD\% (↓)}} &
\rotatebox{80}{\textbf{CR\% (↑)}} & \rotatebox{80}{\textbf{AR\% (↑)}} & \rotatebox{80}{\textbf{AV\% (↓)}} & \rotatebox{80}{\textbf{SR (↑)}} & \rotatebox{80}{\textbf{MDD\% (↓)}} &
\rotatebox{80}{\textbf{CR\% (↑)}} & \rotatebox{80}{\textbf{AR\% (↑)}} & \rotatebox{80}{\textbf{AV\% (↓)}} & \rotatebox{80}{\textbf{SR (↑)}} & \rotatebox{80}{\textbf{MDD\% (↓)}} &
\rotatebox{80}{\textbf{CR\% (↑)}} & \rotatebox{80}{\textbf{AR\% (↑)}} & \rotatebox{80}{\textbf{AV\% (↓)}} & \rotatebox{80}{\textbf{SR (↑)}} & \rotatebox{80}{\textbf{MDD\% (↓)}} \\
\midrule
Buy \& Hold & - &
46.88 & 962.13 & 40.89 & 6.00 & 6.34 &
-6.89 & -35.53 & 29.03 & -1.37 & 15.03 &
0.66 & 4.02 & 30.05 & 0.28 & 11.95 &
18.56 & 176.92 & 71.71 & 1.77 & 20.23 \\
\midrule

\multirow{6}{*}{InvestorAgent}
& GPT-4o            & 33.01 & \textbf{783.38} & 42.32 & 5.37 & 6.34  & -5.33 & -25.93 & 26.99 & -0.98 & 14.36 & -1.22 & -7.10 & 20.77 & -0.25 & 8.79 & -1.98 & -11.29 & 55.58 & 0.06 & 22.13 \\
& GPT-4.1           & \textbf{40.83} & 764.39 & 34.35 & \textbf{6.47} & \textbf{4.38}  & 3.79  & 23.19  & 24.33 & 0.98  & 7.35  & -3.34 & -18.40 & 20.32 & -0.90 & 9.92 & 1.35  & 8.38  & 50.73 & 0.41 & 23.76 \\
& Gemini-2.0-flash  & 10.57 & 80.22  & 42.30 & 1.60 & 8.50  & -13.88& -53.62 & 25.23 & -2.92 & 19.57 & -2.78 & -15.51 & 19.58 & -0.76 & 10.20& -6.39 & -32.64& 64.98 & -0.28& 32.27 \\
& Claude-3.5-haiku  & 11.55 & 124.83 & 38.35 & 2.30 & 8.35  & -8.35 & -41.48 & 25.49 & -1.97 & 15.24 & -7.06 & -35.46 & 30.02 & -1.31 & 15.18& 5.05  & 34.29 & 71.94 & 0.76 & 22.98 \\
& Claude-sonnet-4   & 28.91 & 328.14 & 41.01 & 3.76 & 8.41  & -6.89 & -34.58 & 28.34 & -1.36 & 15.12 & -6.48 & -33.04 & 25.93 & -1.42 & 12.62& 5.56  & 38.20 & 61.40 & 0.83 & 27.17 \\
& \textbf{Vote}     & 29.13 & 293.80 & 37.68 & 3.83 & 8.37  & -5.47 & -28.08 & 27.50 & -1.06 & 13.64 & -2.19 & -12.43 & 22.16 & -0.49 & 9.68 & -1.97 & -11.20 & 58.80 & 0.09 & 28.80 \\
\midrule

\multirow{6}{*}{TradeAgent}
& GPT-4o            & -4.25 & -54.22 & 22.36 & -3.38 & 5.73  & -1.95 & -22.97 & 21.56 & -1.11 & 7.81  & -5.05 & -26.67 & 7.50  & -4.09 & 5.05 & -6.23 & -31.95 & \textbf{31.78} & -1.05& 15.95 \\
& GPT-4.1           & -38.72& -94.33 & 50.68 & -5.38 & 38.72& 4.84  & 27.51  & 23.39 & 1.15  & 6.20  & -5.65 & -29.41 & 28.83 & -1.07 & 9.76 & -11.42& -51.61 & 68.44 & -0.72& 24.75 \\
& Gemini-2.0-flash  & 21.91 & 203.20 & 30.23 & 3.82 & 6.41  & -0.31 & -1.61  & 22.11 & 0.04  & 9.46  & \textbf{5.02}  & \textbf{34.09}  & 24.27 & 1.33  & 5.83 & 9.78  & 74.78 & 52.88 & 1.32 & 16.01 \\
& Claude-3.5-haiku  & -6.62 & -36.48 & 25.72 & -1.64 & 11.05 & -9.45 & -40.61 & 16.81 & 3.01 & 13.85 & -14.51& -60.85 & 24.07 & -3.77 & 16.37& -12.93& -56.31 & 50.12 & -1.40& 15.21 \\
& Claude-sonnet-4   & -5.43 & -24.10 & 26.42 & -0.91 & 12.89 & 14.57 & 121.92 & 21.86 & 3.76  & {3.81} & -13.02& -56.61 & 26.36 & -3.03 & 15.02& -12.35& -54.57 & 41.10 & -1.71& 20.50 \\
& \textbf{Vote}     & -0.36 & -2.30  & 24.44 & 0.02 & 9.03  & 1.22  & 7.55   & 18.90 & 0.48  & 6.70  & -7.28 & -36.39 & 23.83 & -1.78 & 11.53& -9.89 & -46.39 & 40.12 & -1.35& 14.57 \\
\midrule

\multirow{6}{*}{HedgeFundAgent}
& GPT-4o            & -29.15& -84.86 & 40.94 & -4.39 & 29.15 & \textbf{23.70} & \textbf{247.76} & 26.59 & \textbf{4.83}  & 5.59  & -9.09 & -43.46 & 29.97 & -1.76 & 11.14& \textbf{39.66} & \textbf{638.04} & 69.75 & \textbf{3.21} & 14.63 \\
& GPT-4.1           & -29.15& -84.86 & 40.94 & -4.39 & 29.15 & 23.70 & 247.76 & 26.59 & 4.83  & 5.59  & -9.09 & -43.46 & 29.97 & -1.76 & 11.14& 39.66 & 638.04 & 69.75 & 3.21 & 14.63 \\
& Gemini-2.0-flash  & -29.15& -84.86 & 40.94 & -4.39 & 29.15 & 23.70 & 247.76 & 26.59 & 4.83  & 5.59  & -9.09 & -43.46 & 29.97 & -1.76 & 11.14& 39.66 & 638.04 & 69.75 & 3.21 & 14.63 \\
& Claude-3.5-haiku  & -29.15& -84.86 & 40.94 & -4.39 & 29.15 & 23.70 & 247.76 & 26.59 & 4.83  & 5.59  & -9.09 & -43.46 & 29.97 & -1.76 & 11.14& 39.66 & 638.04 & 69.75 & 3.21 & 14.63 \\
& Claude-sonnet-4   & -29.15& -84.86 & 40.94 & -4.39 & 29.15 & 23.70 & 247.76 & 26.59 & 4.83  & 5.59  & -9.09 & -43.46 & 29.97 & -1.76 & 11.14& 39.66 & 638.04 & 69.75 & 3.21 & 14.63 \\
& \textbf{Vote}     & -29.15& -84.86 & 40.94 & -4.39 & 29.15 & 23.70 & 247.76 & 26.59 & 4.83  & 5.59  & -9.09 & -43.46 & 29.97 & -1.76 & 11.14& 39.66 & 638.04 & 69.75 & 3.21 & 14.63 \\
\midrule

\multirow{6}{*}{DeepFundAgent}
& GPT-4o            & 13.89 & 95.19  & 37.14 & 1.98 & 7.61  & 3.72  & 22.68  & 25.89 & 0.92  & 8.92  & -0.30 & -1.80  & 24.35 & 0.04  & 7.98 & 18.03 & 169.68 & 71.07 & 1.74 & 21.86 \\
& GPT-4.1           & -5.52 & -27.77 & 37.30 & -0.69& 19.22 & 4.86  & 29.67  & \textbf{16.67} & 1.64  & \textbf{3.49}  & 0.44  & 2.68   & 1.08  & \textbf{2.45}  & \textbf{0.00} & 2.89  & 18.56  & 58.62 & 0.57 & 26.08 \\
& Gemini-2.0-flash  & -8.04 & -55.62 & \textbf{20.40} & -3.88& 8.04 & 3.59  & 20.84  & 24.77 & 0.89  & 6.57  & 0.00  & 0.00   & \textbf{0.00}  & 0.00  & 0.00 & -13.11& -56.85 & 40.55 & -1.87& 22.20 \\
& Claude-3.5-haiku  & -15.43& -57.76 & 38.49 & -2.04& 28.48 & -1.12 & -6.01  & 26.80 & -0.10 & 12.25 & 1.67  & 10.40  & 27.51 & 0.50  & 7.91 & 10.74 & 84.08  & 69.93 & 1.21 & 23.11 \\
& Claude-sonnet-4   & -6.96 & -32.07 & 35.81 & -0.90& 12.99 & 5.48  & 34.81  & 25.27 & 1.31  & 8.52  & -8.24 & -40.23 & 21.91 & -2.24 & 10.18& 8.20  & 60.29  & 56.87 & 1.11 & 18.58 \\
& \textbf{Vote}     & 8.61  & 55.75  & 36.68 & 1.39 & 10.14 & 9.45  & 52.43  & 22.82 & 1.96  & 6.42  & -0.54 & -3.20  & 1.33  & -2.45 & 0.54 & 4.91  & 33.22  & 65.65 & 0.75 & \textbf{13.00} \\

\bottomrule
\end{tabular}
\end{adjustbox}

\caption{Merged live trading outcomes for four agents across five backbone models on two stock assets (TSLA, BMRN) and two crypto assets (BTC, ETH), from Aug 1 to Sep 30. Metrics: CR (Cumulative Return), AR (Annualized Return), AV (Annualized Volatility), SR (Sharpe Ratio), and MDD (Maximum Drawdown). “Vote” denotes majority-vote ensembles across the backbones.}
\label{tab:agent_model_merged}
\end{threeparttable}
\end{table*}

\subsection{Performance Analytics Interface (PAI)}

Each agent makes one trading decision per day, represented as a signal \( w_t \in \{1, -1, 0\} \), 
where \( w_t = 1 \) means buying the asset, \( w_t = -1 \) means selling it, and \( w_t = 0 \) means holding the current position.  
The agent’s daily return is:
\(
r_t = w_t \frac{P_t - P_{t-1}}{P_{t-1}},
\)
and its overall performance over time is tracked through cumulative return:
\(
R_t = \prod_{i=1}^{t} (1 + r_i) - 1.
\)

To assess trading performance, we employ five widely used financial indicators.  
The \textbf{Cumulative Return (CR)}~\cite{hull2012risk} measures the total growth of an agent’s portfolio over the evaluation period, capturing its overall profitability.  
The \textbf{Annualized Return (AR)}~\cite{sharpe1966mutual} adjusts this value to a yearly rate, \(
 R_{\text{annual}} = (1 + R_t)^{\frac{252}{T}} - 1
\)
allowing comparisons across different time spans.  
To quantify risk, we compute the \textbf{Annualized Volatility (AV)}~\cite{hull2018options} as \( \sigma_{\text{annual}} = \sqrt{252} \times \text{std}(r_t) \), which reflects how much the agent’s daily returns fluctuate.  
The \textbf{Sharpe Ratio (SR)}~\cite{sharpe1998sharpe}, given by \( S = \frac{\bar{r} - r_f}{\sigma_{\text{annual}}} \), evaluates how efficiently the agent converts risk into return, assuming a risk-free rate \( r_f = 0 \).  
Finally, the \textbf{Maximum Drawdown (MDD)}~\cite{magdon2004maximum} is defined as \( D_{\max} = \max_t \left( \frac{\max_{i \leq t} R_i - R_t}{\max_{i \leq t} R_i} \right) \), capturing the largest observed loss from a peak to a trough and indicating downside exposure.

All metrics are updated daily and displayed on a real-time leaderboard that compares the profitability, stability, and adaptability of agents across different LLM backbones and market types.  

\section{Experiment Settings}

\textbf{LLM selection.} For each agent, we evaluate five different backbone models from OpenAI, Claude, and Gemini. Specifically, we include GPT-4o, GPT-4.1, Gemini-2.0-flash, Claude-3.5-haiku, and Claude-sonnet-4.

\noindent\textbf{Assets.} We evaluate both equity and cryptocurrency markets to capture distinct market dynamics and behavioral patterns. Our representative test assets include Bitcoin (BTC) and Ethereum (ETH) from the cryptocurrency space, and two publicly traded stocks, Tesla (TSLA) and BioMarin Pharmaceutical (BMRN), representing the high-technology and biomedicine sectors, respectively.

Cryptocurrencies such as BTC and ETH operate in highly volatile, sentiment-driven environments that react rapidly to macroeconomic and policy signals, providing a rigorous test of model adaptability under extreme market fluctuations. \cite{buthelezi2025cryptocurrency} In contrast, equities like TSLA and BMRN are influenced by sector-specific fundamentals—technological innovation, production metrics, or clinical trial results—allowing us to assess how well the system handles structured, information-rich contexts. \cite{richardson2012makes} By combining these heterogeneous assets, we ensure that the evaluation spans diverse asset classes, time sensitivities, offering a comprehensive benchmark for decision-making agents in real-world trading conditions.

\noindent\textbf{Agent initialization and evaluation.} To establish historical grounding and construct contextual memory, agents undergo an initialization phase with data from 2025-05-01 to 2025-07-31, during which they adapt to recent market dynamics, simulate realistic trading environments, and form representative portfolio positions. Real-time evaluation begins on 2025-08-01. As of the paper’s drafting date: 2025-09-30, the agents have completed two months of live trading and continue to operate autonomously, producing daily trading actions and performance evaluations in real time.



\section{Results}

\begin{figure*}[t]
\centering
\begin{subfigure}{0.45\textwidth}
\centering
\includegraphics[width=\linewidth]{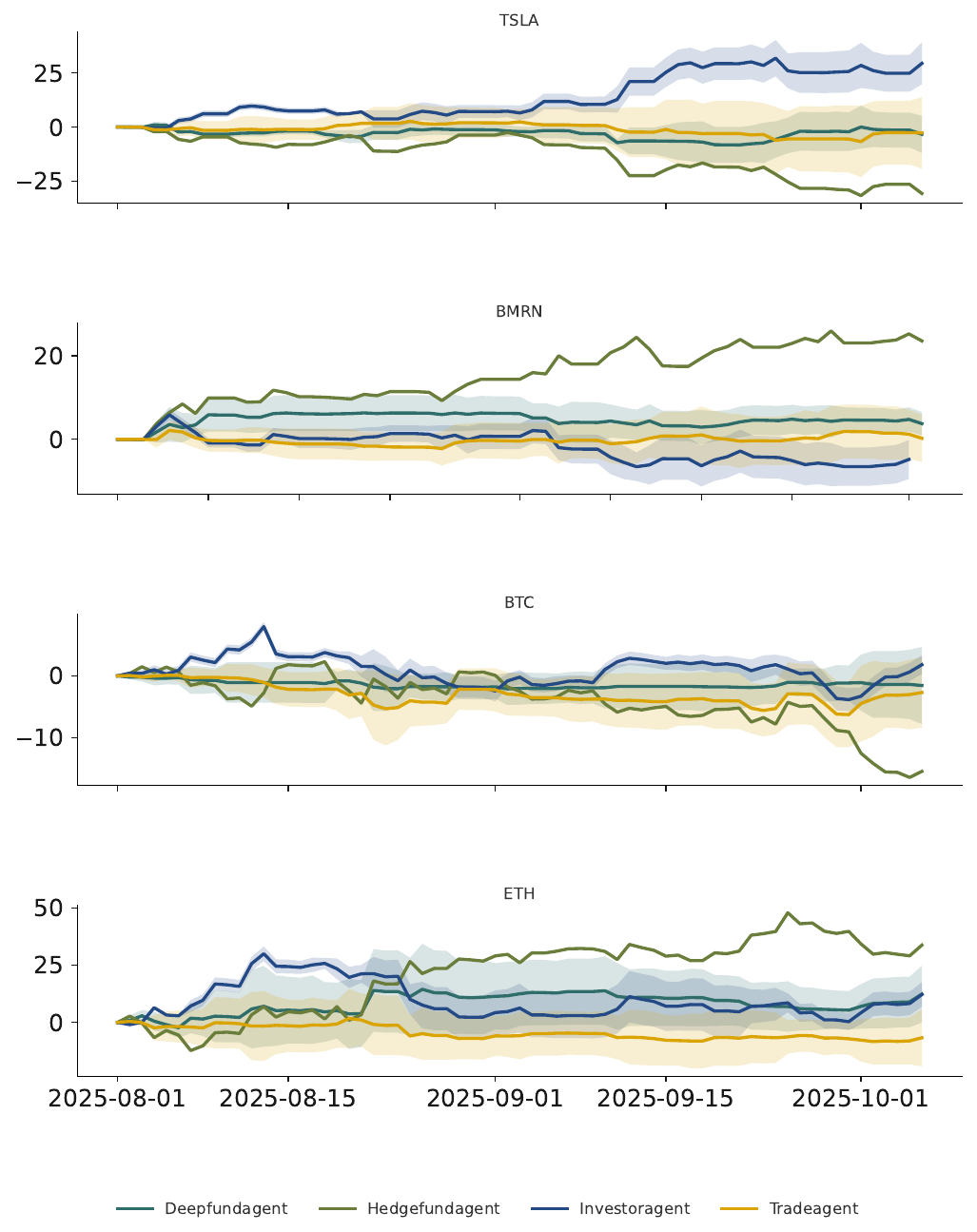}
\caption{Comparison of aggregated performance of different agents across four assets.}
\label{fig:cr_btc}
\end{subfigure}\hfill
\begin{subfigure}{0.45\textwidth}
\centering
\includegraphics[width=\linewidth]{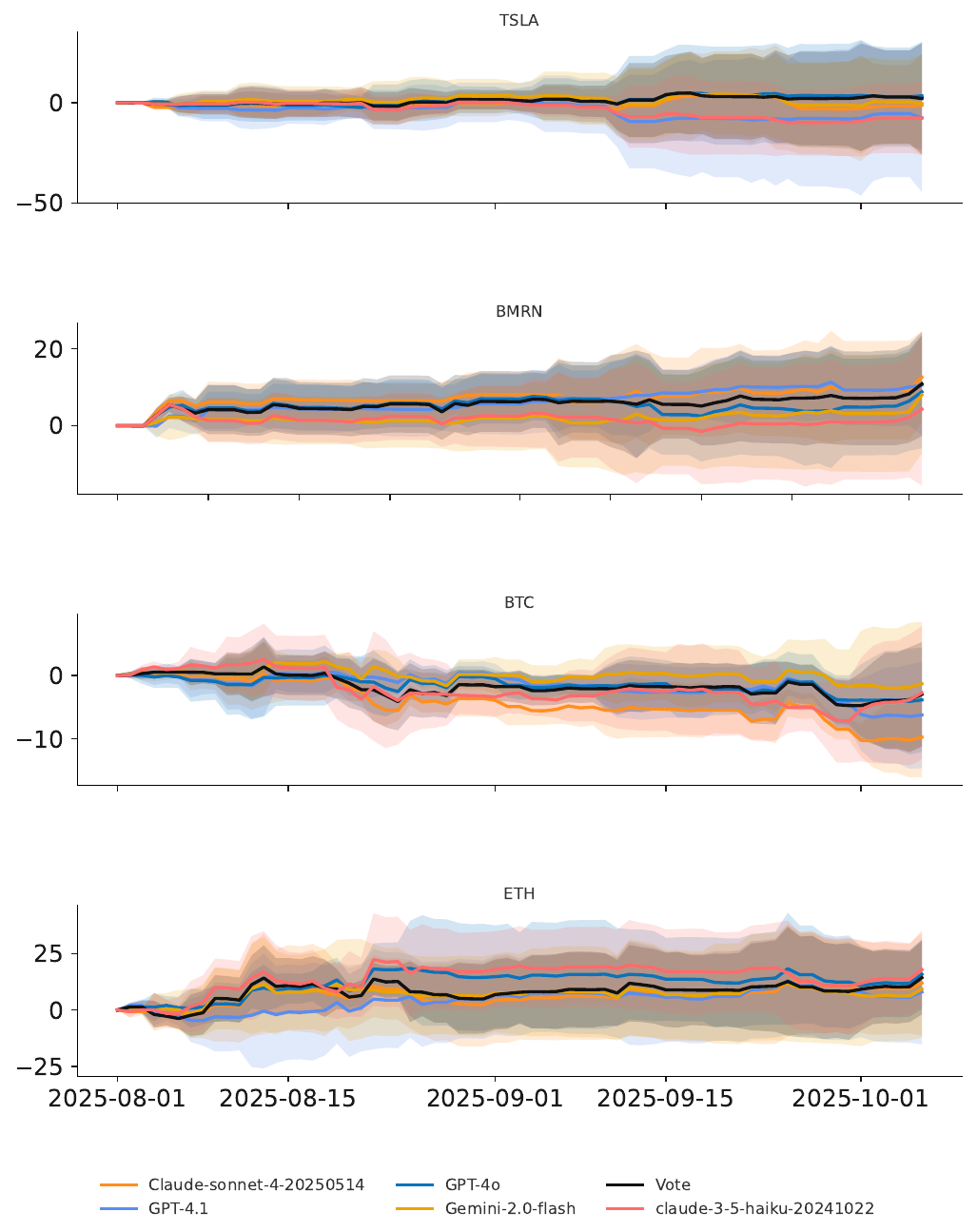}
\caption{Comparison of aggregated performance of different LLMs across four assets. }
\label{fig:cr_tsla}
\end{subfigure}
\caption{Aggregated performance of different agents and LLMs across four assets. The solid line represents the average cumulative return, while the shaded area indicates the range between the maximum and minimum CR observed when switching LLMs or agent frameworks.}
\label{fig:cr_btc_tsla}
\end{figure*}



In this section, we examine how different agents perform in a continuously evolving live trading environment 
to address the following research questions (RQs):
\begin{itemize}
    \item \textbf{RQ1}: Can LLM-based agents truly trade, meaning make profitable and consistent decisions that outperform simple Buy \& Hold strategies in live markets?
    \item \textbf{RQ2}: Which factor contributes more to trading performance, the agent’s structural framework or the reasoning ability of its underlying LLM backbone?
    \item \textbf{RQ3}: How effectively can agents interpret and translate complex market information into trading decisions that anticipate and adapt to price fluctuations?
    \item \textbf{RQ4}: How do different trading philosophies, from aggressive to conservative strategies, influence decision-making behavior and overall profitability across markets?

\end{itemize}

\subsection{Performance of Agents in Live Trading (RQ1)}

Table~\ref{tab:agent_model_merged} summarizes the live trading results of all agents across both cryptocurrency and stock markets. Overall, the findings confirm that \textbf{LLM-based agents can indeed trade profitably in real time}, often surpassing simple buy-and-hold strategies while maintaining stability across dynamic market conditions. The outcomes, however, vary significantly across agents and asset types, illustrating that performance depends heavily on design philosophy and market context.

Among the evaluated systems, \textbf{DeepFundAgent} delivers the most balanced and consistent results, achieving cumulative returns of 8.61\% on TSLA and 9.45\% on BMRN, with annualized volatilities of 36.68\% and 22.82\% and Sharpe ratios of 1.39 and 1.96, respectively. \textbf{InvestorAgent} also performs competitively, particularly under GPT-4.1, reaching 40.83\% cumulative return with a Sharpe ratio of 6.47 on TSLA. It also shows sensitivity to backbone models and assets, with its CR on BMRN dropping from 3.79\% (GPT-4.1) to –6.89\% (Claude-sonnet-4). \textbf{TradeAgent} and \textbf{HedgeFundAgent} display wider fluctuations, with TradeAgent’s returns on TSLA ranging from –38.72\% (GPT-4.1) to 21.91\% (Gemini-2.0-Flash). HedgeFundAgent demonstrates mixed outcomes, achieving 39.66\% CR on ETH in one setup but negative returns on BTC. Yet it maintains relative robustness across backbones due to its hierarchical coordination with sixteen specialized sub-agents feed two managerial agents that aggregate signals, reducing variance introduced by individual LLMs.

\begin{figure*}
\centering
\centering
\includegraphics[width=0.7\textwidth]{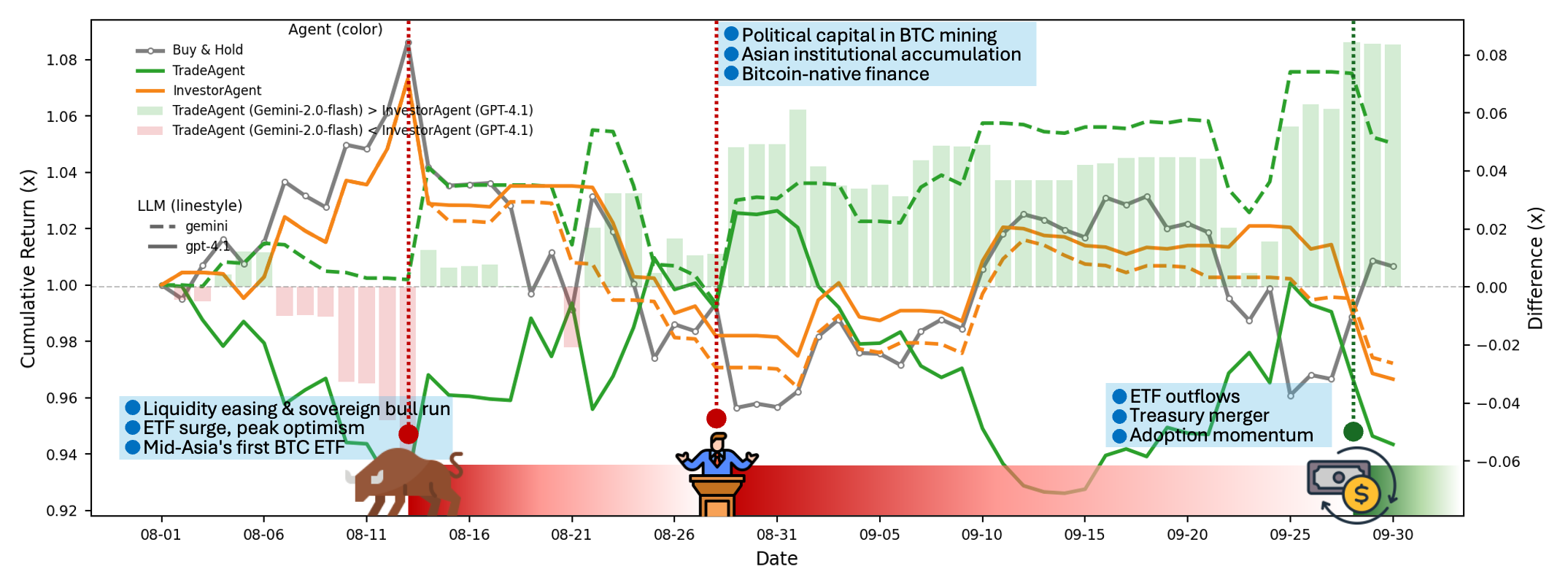}
\caption{Agent performance on BTC under different market events. The bars represent the profit gap between TradeAgent (Gemini-2.0-flash) and InvestorAgent (GPT-4.1).}
\label{fig:case}

\end{figure*}

Across these experiments, no single LLM or agent consistently outperforms others across all assets. Nevertheless, the results reveal clear evidence of agent–model complementarity, where certain pairings yield notably stronger outcomes. For instance, TradeAgent coupled with Gemini-2.0-flash attains leading performance on TSLA, ETH, and BTC, and ranks second on BMRN. InvestorAgent performs best when paired with GPT-4.1, delivering strong results on both TSLA and BMRN. Similarly, DeepFundAgent achieves its highest returns with GPT-4o, exhibiting robust performance on ETH and TSLA and maintaining competitiveness across remaining assets. This underscores the importance of conducting multi-market evaluations, which offer broader insights compared to analyses restricted to a limited set of assets and results.


\subsection{Which Matters More: Agent Architecture or LLM Backbone? (RQ2)}

To disentangle the relative influence of agent design and LLM backbone on live trading performance, we aggregate outcomes across all agent–model pairings, as illustrated in Figure~\ref{fig:cr_btc_tsla}. Figure \ref{fig:cr_btc} reports the performance of each agent under different LLMs, while Figure \ref{fig:cr_tsla} summarizes how each LLM performs across agent architectures.

Overall, the results reveal that modifying the underlying LLM within a fixed agent framework produces only modest changes in profitability, whereas varying the agent design while holding the LLM constant leads to far greater performance divergence. For instance, in Figure \ref{fig:cr_btc}, the InvestorAgent results on TSLA cluster tightly, showing minimal spread across LLM configurations, indicating that switching models has limited effect. By contrast, Figure \ref{fig:cr_tsla} shows GPT-4.1 (light blue) spanning a wide range of returns across agents, demonstrating that the structural design of the agent exerts a stronger influence on trading outcomes than the reasoning capability of the backbone itself.

These findings suggest that \textbf{an agent’s architecture, including its decision logic, coordination strategy, and risk control mechanisms, plays a more decisive role in shaping profitability and stability} than merely upgrading to a more powerful model. Within the \textsc{\ama} environment, enhancing the way agents utilize LLM may thus yield greater gains than improving the LLMs alone.
\subsection{Understanding Agents' Decision-Making Driven by Market Signals (RQ3)}

To better understand how agents make trading decisions under real-time and volatile market conditions, we analyzed the behavior of \textbf{TradeAgent} and \textbf{InvestorAgent} on BTC, using \textbf{Gemini-2.0-flash} and \textbf{GPT-4.1} as backbones. BTC was selected for its high sensitivity to market news and sentiment. We compared their cumulative return trajectories against the Buy \& Hold baseline (Figure~\ref{fig:case}) and examined how each agent interpreted macroeconomic signals and adapted its strategy during major volatility events.
We focus on three key episodes: August 13, August 28, and September 28, which capture both successful trades and failed forecasts.

On \textbf{August 13–14}, global equity markets entered a period of broad-based bullish momentum, with major indices across the U.S., Europe, and Asia reaching record highs. This synchronized upswing, fueled by policy-easing expectations and renewed sovereign inflows, was widely described by media as one of the most coordinated global rallies in recent years.~\cite{Oguh2025_GlobalMarkets13Aug} Despite overwhelmingly bullish sentiment, TradeAgent identified the fragility of this extreme optimism, recognizing that an “unprecedented” event lacked historical equilibrium. It correctly hedged risk near the peak, profiting when prices normalized.  
On \textbf{August 28–29}, During August 28–29, several rare structural shifts occurred:
(1) Political capital met mining, as the Trump-backed American Bitcoin Corp. listed on NASDAQ via its merger with Gryphon Digital Mining, marking one of the first clear intersections between politics and Bitcoin infrastructure.
(2) Institutional accumulation surged in Asia, with major Japanese and Korean firms announcing plans to integrate Bitcoin into their fiscal reserves.
(3) Ecosystem innovation advanced, driven by the RGB protocol and Tether’s integration on the Bitcoin network, expanding Bitcoin’s role beyond a store of value.
Although these were long-term bullish signals, TradeAgent detected short-term exhaustion in technicals and executed a SELL, accurately anticipating a short-term correction.
By contrast, on \textbf{September 28–29}, the agent’s short position, initially justified by ETF outflows and synchronized drawdowns, was reversed by a sudden surge in liquidity and sovereign inflows, leading to a short squeeze. All agents shared this misjudgment, reflecting the difficulty of reacting to abrupt macro reversals.

\begin{figure*}
\centering
\centering
\includegraphics[width=0.7\textwidth]{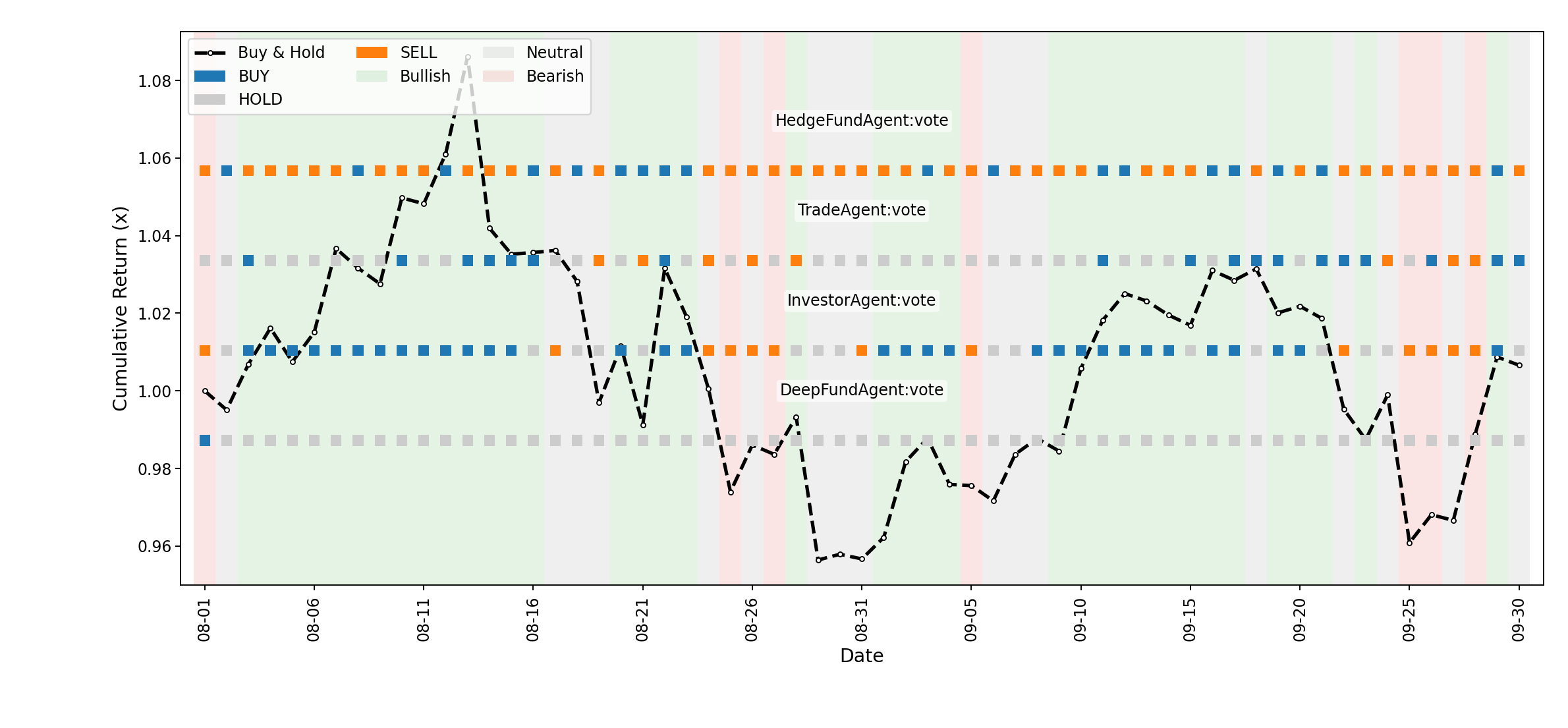}
\caption{Cumulative return comparison of Buy-and-Hold baseline for BTC (Aug–Sep 2025), annotated with daily news sentiment (green = bullish, gray = neutral, red = bearish) and agent voting signals (squares: blue = BUY, gray = HOLD, orange = SELL.}
\label{fig:sentiment}

\end{figure*}


Figure~\ref{fig:sentiment} shows that for most of the period, the performance gap between InvestorAgent (GPT-4.1) and TradeAgent (Gemini-2.0-flash) remained narrow. InvestorAgent captured the early rally around August 13 but missed later turning points, allowing TradeAgent to overtake it by capitalizing on volatility and adapting more dynamically to shifting sentiment. These results suggest that \textbf{the most successful agents are those capable of exploiting periods of volatility rather than merely following long-term trends}, as agents that take advantages of volatility fits more for the daily trading frequency of \textsc{\ama}.

This analysis also highlights the value of a continuously updating, multi-class-asset benchmark built on verified and diverse market data. Real-time developments across both equities and cryptocurrencies demonstrate that agent performance cannot be fairly assessed through static or single-market evaluations. By integrating verified data streams and standardized decision protocols, \textsc{\ama} enables systematic comparison of multiple agents under consistent and dynamic market conditions. This design provides a more comprehensive and generalizable understanding of how different agent architectures interpret across real-world financial environments.



\subsection{Revealing the Influence of Agent Trading Styles and Patterns (RQ4)}

We further investigate how trading styles shape agent behavior in the \textsc{\ama} framework. Figure~\ref{fig:sentiment} shows the interplay between daily market sentiment, BTC price dynamics, and agent voting behaviors.

Although all agents operate on identical market information, their trading strategies diverge from the first day. HedgeFundAgent and InvestorAgent favored short positions, whereas DeepFundAgent preferred buying and TradeAgent tended to hold. These strategic contrasts became more distinct as trading progressed, reflecting each framework’s inherent risk preferences and decision logic.

Across the two-month live trading period, overall market sentiment remained optimistic. Nevertheless, HedgeFundAgent persistently maintained short positions, revealing its contrarian and high-risk nature. TradeAgent, despite being generally more volatile, chose to hold positions on most days, indicating that while inherently aggressive, it could exhibit conservative behavior depending on its backbone model. InvestorAgent’s decisions largely followed prevailing market trends, as reflected in its cumulative return curve (Figure~\ref{fig:cr_btc}), which outperformed others for most of the trading period. The agent frequently executed buy and sell actions, exhibiting a proactive and risk-taking trading style. Nevertheless, its overall profitability remained moderate across assets, suggesting that it adopted a relatively cautious stance during periods of elevated risk.
In contrast, DeepFundAgent demonstrated a more conservative decision pattern, frequently opting for the “vote” action and maintaining stable returns across multiple assets. The prudence observed in both agents’ strategies may stem from their memory-adaptive mechanisms, which enable them to draw on prior high-risk experiences and adjust decisions toward more risk-averse behaviors.

These findings, consistent with Table~\ref{tab:agent_model_merged}, underscore that \textbf{trading style is a key determinant of profitability and risk exposure}. DeepFundAgent achieved consistent, moderate gains, while the aggressive HedgeFundAgent generated substantial profits in ETH  and BMRN, which are 39.66\% and 23.70\% respectively, but incurred significant losses in TSLA and BTC. Together, these results reaffirm a fundamental market principle, higher risk can lead to greater reward but also greater potential loss.

\section{Conclusion}

In this work, we introduced \textsc{\ama}, the first continuously updating, multi class-asset benchmark for evaluating LLM-based trading agents in real time. Built upon three core components, including verified multi-market information streams, a unified trading protocol, and a transparent performance interface, \textsc{\ama} establishes a reproducible and dynamic foundation for studying financial trading agents under genuine market conditions. It bridges the gap between language understanding and live decision-making.

Our findings reveal four key insights. First, LLM-based agents can trade effectively and consistently outperform simple buy \& hold strategies in live environments. Second, agent architecture, rather than the choice of LLM backbone, exerts the strongest influence on profitability and adaptability. Third, agents demonstrate distinct reasoning behaviors when interpreting complex, fast-changing market signals, with more adaptive frameworks proving better at navigating volatility. Finally, trading styles, ranging from aggressive to conservative, shape each agent’s risk–return profile and stability over time. In future work, we plan to extend \textsc{\ama} with inter-agent communication, cross-asset dynamics, and reinforcement learning feedback, advancing toward a deeper understanding of autonomous financial intelligence in LLM-driven systems.



\clearpage


\bibliographystyle{ACM-Reference-Format}
\bibliography{sample-base}

\clearpage
\appendix
\section{Details of quality control}
\label{quality}

To ensure the daily market information delivered to agents is of high quality, we conducted multiple iterative refinements of the summarization prompt. Initially, we performed a preliminary review and adjustment based on data collected from May 1 to May 10, which significantly improved generation quality. Subsequently, we sampled 20 days of BTC and TSLA news from cryptocurrency and stock markets, respectively, and had two domain experts independently verify their accuracy and coverage to ensure the quality and reliability of the generated summaries.

This iterative process ensured that the final summaries captured both market breadth and factual precision, minimizing redundancy and bias across sources. As a result, the refined daily inputs provided agents with consistent, verified, and information-rich market contexts for decision-making.

\begin{tcolorbox}[colback=gray!15,colframe=gray!60,title={\textbf{Quality Control Guideline}},sharp corners,boxrule=0.5pt]

\textbf{Purpose.} This guideline outlines the procedures to ensure high-quality and unbiased daily news summaries used by agents.

\vspace{0.5em}
\textbf{1. Objective}

Daily summarized news must accurately represent the key information from multiple sources without introducing bias. Each asset’s daily summary should be:
\begin{itemize}
\item \textbf{Accurate}: Correctly aligned with the trading date.
\item \textbf{Comprehensive}: Cover all major price-driving categories.
\item \textbf{Unbiased}: Maintain balanced sentiment and tone.
\item \textbf{Diverse}: Incorporate multiple independent sources.
\end{itemize}

\vspace{0.5em}
\textbf{2. Rules and Scoring}

\textbf{Rule 1: Date Accuracy} – Ensure all articles correspond to the correct trading date.
\begin{itemize}
\item 0 = Not aligned – multiple off-date items.
\item 1 = Mostly aligned – minor off-date presence.
\item 2 = Fully aligned – all sources match the target date.
\end{itemize}

\textbf{Rule 2: Coverage of Key News Drivers} – Include all major asset-specific developments.
\begin{itemize}
\item 0 = Very limited – one category only.
\item 1 = Partial – two to three categories.
\item 2 = Comprehensive – all four key drivers.
\end{itemize}

\textbf{Rule 3: Bias Awareness} – Evaluate sentiment balance across sources and summaries.
\begin{itemize}
\item 0 = Skewed – one-sided sentiment.
\item 1 = Moderate – partial balance.
\item 2 = Balanced – neutral overall tone.
\end{itemize}

\textbf{Rule 4: Source Diversity} – Verify multi-outlet representation.
\begin{itemize}
\item 0 = Low diversity – one outlet dominant.
\item 1 = Moderate – two to three outlets.
\item 2 = High diversity – four or more outlets.
\end{itemize}

\vspace{0.5em}


\end{tcolorbox}

\begin{tcolorbox}[colback=gray!15,colframe=gray!60,title={\textbf{Prompt For Summarization}},sharp corners,boxrule=0.5pt]

\textbf{Task:} Analyze the provided \{symbol\} news articles from \{date\_str\} and produce a clear, comprehensive, and well-structured summary.

\vspace{0.5em}
\textbf{Instructions:}
\begin{itemize}
\item Use \textbf{only} the provided articles for your analysis.
\item Do \textbf{not} retrieve or reference any external information.
\item Refrain from including current prices or speculative forecasts.
\item Focus strictly on the \textbf{events, sentiment, and contextual details} presented in the articles.
\item Do not mention or enumerate article IDs in the output.
\end{itemize}

\vspace{0.5em}
\textbf{Articles from \{date\_str\}:}

\{articles\_text\}

\vspace{0.5em}
\textbf{Output Requirements:}
\begin{enumerate}
\item Provide a cohesive summary highlighting the key developments for \{symbol\}.
\item Identify major themes or emerging narratives within the content.
\item Assess and articulate the overall market sentiment reflected in the articles.
\end{enumerate}

Compose your response as a coherent and polished narrative that maintains logical flow and clarity throughout.

\end{tcolorbox}

\section{Details of agent execution protocol}
\label{parameter}

\begin{tcolorbox}[colback=gray!15,colframe=gray!60,title={\textbf{Prompt For Summarization}},sharp corners,boxrule=0.5pt]


\vspace{0.5em}
You are a professional financial decision-making agent specialized in quantitative and fundamental reasoning with a daily trading frequency.  
Your primary task is to analyze the reasoning outputs of other agent roles and integrate their insights into a unified, evidence-based conclusion.  

Based on your analysis and the provided definitions, determine the most appropriate trading decision for the target asset \textbf{[ASSET]} given its current market price \textbf{[PRICES]}and contextual signals.  

Your possible actions are defined as follows: 

- \textbf{Buy}: Indicates a bullish outlook or perceived undervaluation, suggesting the asset price is likely to rise. And you choose to be in long position.

- \textbf{Sell}: Indicates a bearish outlook or perceived overvaluation, suggesting the asset price is likely to fall. And you choose to be in short position.  

- \textbf{Hold}: Indicates market uncertainty or equilibrium, suggesting no immediate trading action. And you choose to go flat position.

Return your final output strictly in the following format:

[Decision]: Buy / Sell / Hold

\end{tcolorbox}

\textbf{System Prompt Definition.}
Each agent operates under a structured system prompt that defines its analytical objective and reasoning behavior. The prompt guides the model to integrate multi-source financial information and produce consistent, interpretable trading decisions. Specifically, the agent is assigned the following role and task. This design ensures interpretability and reproducibility across agents and backbones.  
For consistency and fairness, all LLM-based agents share identical generation settings, including temperature, top-$p$, and maximum token length, as detailed below.

\textbf{Hyperparameters.}
Table~\ref{tab:deepfund_params} summarizes the key parameters used in DeepFund. These parameters control model randomness, retry behaviors, data coverage, and decision memory. All settings remain fixed during evaluation to ensure a fair comparison across models and runs.

\begin{table*}[h!]
\centering
\begin{tabular}{lcl}
\toprule
\textbf{Parameter} & \textbf{Default Value} & \textbf{Usage} \\
\midrule
LLM temperature & 0.5 & Controls randomness of the LLM inference. \\
Retry times & 3 & Number of retries for LLM inference if the LLM do not give correct response. \\
Warm-up windows & 90 & Trading days that used to initialize the agents. \\
Decision memory size & 7 & Number of past recent actions and price history for decision-making. \\
\bottomrule
\end{tabular}
\caption{DeepFund parameter settings.}
\label{tab:deepfund_params}
\end{table*}

\section{Performance analysis interface}
\label{interface}

\begin{figure*}[t]
    \centering
    \includegraphics[width=0.6\linewidth]{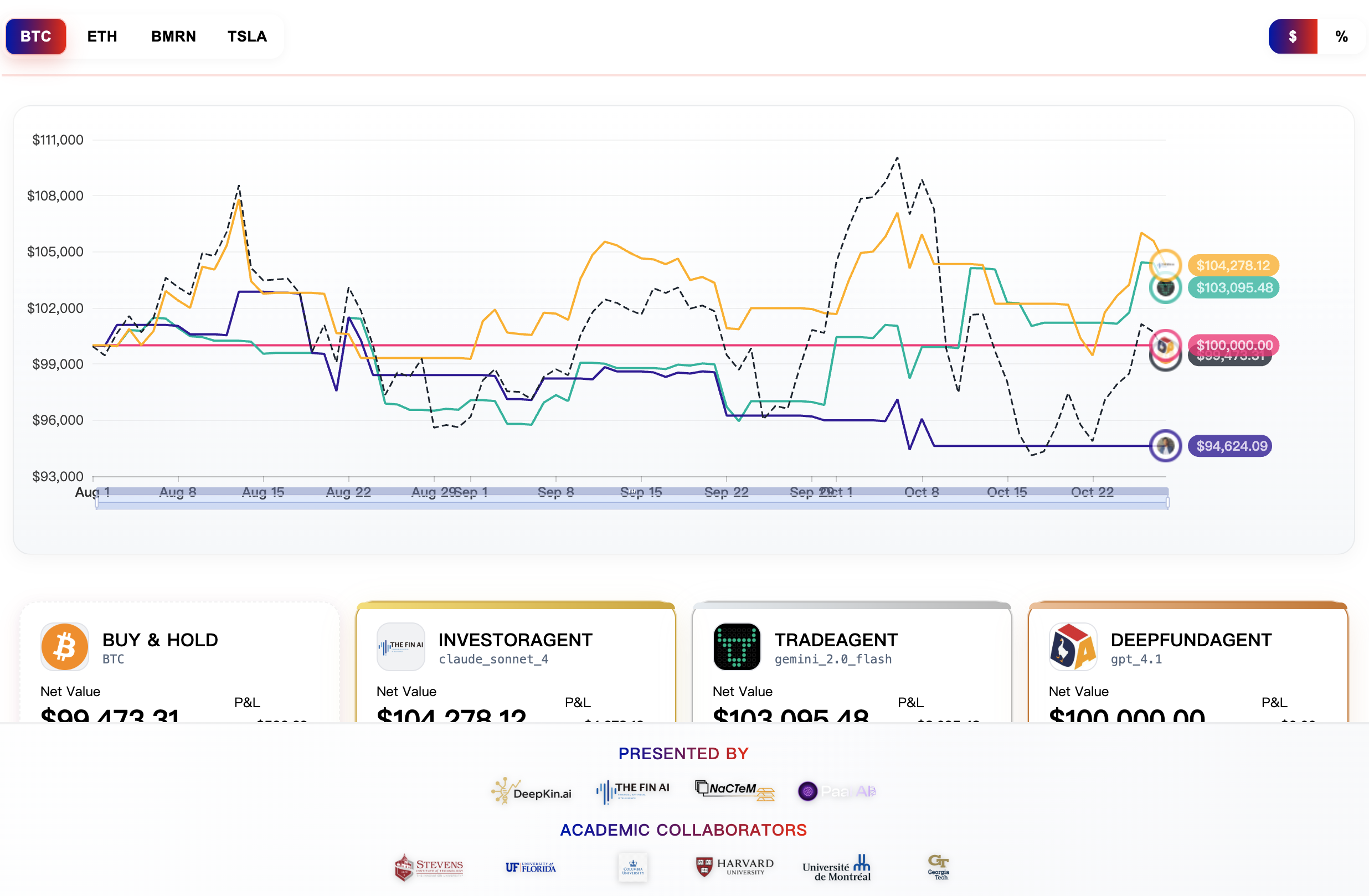}
    \caption{\textbf{Agent Market Arena Dashboard (Overview View).} The interface provides a unified view for monitoring, comparing, and analyzing the real-time performance of LLM-based trading agents across assets, models, and strategies.}
    \label{fig:interface1}
\end{figure*}

\begin{figure*}[t]
    \centering
    \includegraphics[width=0.6\linewidth]{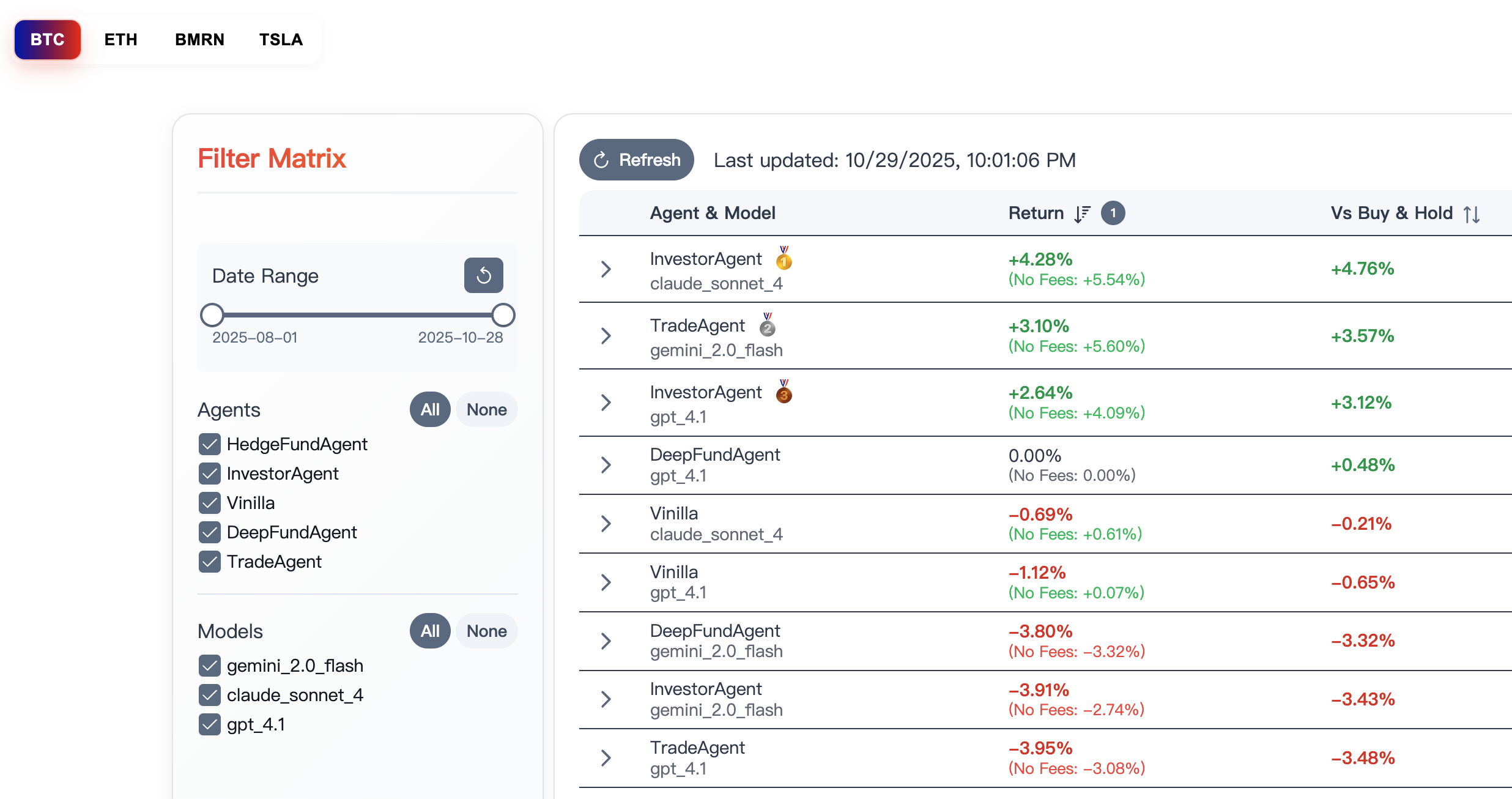}
    \caption{\textbf{Equity Comparison View.} Users can select specific agents, assets, and models to visualize comparative equity curves, enabling multi-dimensional evaluation of performance dynamics.}
    \label{fig:interface2}
\end{figure*}

To enable transparent and reproducible evaluation of live trading behaviors, we built an interactive web-based monitoring interface (Figures~\ref{fig:interface1}–\ref{fig:interface2}) for \textsc{\ama}. This platform allows users to observe, analyze, and compare the performance of LLM-driven trading agents in real time.

\paragraph{Overview Dashboard.}
The main interface (\autoref{fig:interface1}) aggregates all agent performances under live market conditions. It displays core metrics including total balance, cumulative return (with and without fees), Sharpe ratio, and win rate. Agents are dynamically ranked by returns, while an automatic refresh mechanism (default: every 30 seconds) ensures synchronization with real-time market data streams. 
Users can quickly identify outperforming agents and drill down for detailed performance trajectories.

The filtering panel supports four orthogonal dimensions:
\begin{itemize}
    \item \textbf{Agents:} \texttt{TradeAgent}, \texttt{InvestorAgent}, \texttt{HedgeFundAgent}, and \texttt{DeepFundAgent}.
    \item \textbf{Assets:} e.g., \texttt{BTC}, \texttt{ETH}, \texttt{TSLA}, and \texttt{BMRN}, covering both financial and crypto markets.
    \item \textbf{Models:} backbones including \texttt{GPT-4.1}, \texttt{GPT-4o}, et,al. 
    \item \textbf{Strategies:} such as \texttt{Baseline}, \texttt{Flip on Reversal}, and \texttt{Pyramid Exit}, supporting customized evaluation modes.
\end{itemize}
This multi-axis filtering framework allows flexible comparison across model architectures, market domains, and trading strategies.

\paragraph{Equity Comparison View.}
The second interface (\autoref{fig:interface2}) introduces an interactive visualization module for cross-agent equity analysis. Users can select subsets of agents, assets, and LLM models to render synchronized \emph{equity curves} that represent portfolio growth trajectories over time. Each line denotes one agent-model combination (e.g., \texttt{InvestorAgent-BTC-GPT4o}), with dynamic tooltips revealing daily balance, model type, and strategy.

This design enables detailed visual inspection of relative performance dynamics, stability, and divergence patterns under identical market signals. Researchers can thus examine how agents differ in volatility tolerance, recovery behavior, and decision consistency, offering an intuitive complement to tabular metrics.

\paragraph{Analytical Utility.}
Together, the two interfaces provide both a \emph{macro-level} snapshot and a \emph{micro-level} analytical view:
\begin{itemize}
    \item The \textbf{Overview Dashboard} facilitates longitudinal tracking and ranking of live agents.
    \item The \textbf{Equity Comparison View} supports temporal and inter-agent analysis for understanding decision-making robustness.
\end{itemize}

By integrating these views, the system allows researchers to systematically assess the interplay between agent architectures, LLM backbones, and strategy adaptability under continuously evolving market environments.

\end{document}